\documentclass[journal]{IEEEtran}

\usepackage{times}
\usepackage{epsfig}
\usepackage{graphicx}
\usepackage{subfigure}
\usepackage{amsmath}
\usepackage{amsfonts}
\usepackage{amssymb}
\usepackage{booktabs}
\usepackage{algorithm}
\usepackage{algorithmic}
\usepackage{multirow}

\usepackage{diagbox}
\usepackage{rotating}
\usepackage{color}
\usepackage{bm}
\usepackage{caption}

\usepackage{enumerate}

\usepackage{pifont}
\usepackage{overpic}
\usepackage{enumitem}

\newcommand{\et}{\textit{et al}}
\newcommand{\ie}{\textit{i.e.}}%
\newcommand{\etc}{\textit{etc}}
\newcommand{\wrt}{\textrm{w.r.t}}
\definecolor{pp}{rgb}{0.827,0, 0.658}
\newcommand{\cmark}{\ding{51}}%
\newcommand{\xmark}{\ding{55}}%

\usepackage[pagebackref=true,breaklinks=true,colorlinks,bookmarks=false]{hyperref}

\ifCLASSINFOpdf
\else
\fi
\begin{document}

\title{A Universal Model for Cross Modality Mapping \ \ \  
by Relational Reasoning}

\author{Zun Li,
        Congyan Lang,
        Liqian Liang,
        Tao Wang,
        Songhe Feng,
        Jun Wu,
        and
        Yidong Li
        % <-this % stops a space
\thanks{Zun Li, Congyan Lang, Liqian Liang, Tao Wang, Songhe Feng, Jun Wu and Yidong Li are with the School of Computer and Information Technology at Beijing Jiaotong University, Beijing, 100044, 
China. E-mail:(zlee2016@bjtu.edu.cn, cylang@bjtu.edu.cn, lqliang@bjtu.edu.cn, twang@bjtu.edu.cn, shfeng@bjtu.edu.cn, wuj@bjtu.edu.cn
and ydli@bjtu.edu.cn).
Congyan Lang is the corresponding author.}

%\thanks{Submitted Manuscript for TNNLS Special Issue on Deep Neural Networks for Graphs: Theory, Models, Algorithms and Applications }
}

\markboth{Journal of \LaTeX\ Class Files,~Vol.~14, No.~8, February~2021}%
{Shell \MakeLowercase{\textit{et al.}}: Bare Demo of IEEEtran.cls for IEEE Journals}
% The only time the second header will appear is for the odd 

\maketitle

\begin{abstract}
With the aim of matching a pair of instances from two different modalities,
cross modality mapping has attracted growing attention in the computer vision community.
Existing methods usually formulate the mapping function as the similarity measure between the pair of instance features,
which are embedded to a common space.
%where intra and inter relations in single and cross modalities are not well explored.
%In this paper,
However,
we observe that the relationships among the instances within a single modality (intra relations) and those between the pair of heterogeneous instances (inter relations) 
are insufficiently explored in previous approaches.
%and such intrinsic relations can be modeled as a structural and reasoning relationship, respectively.
%Motivated by this, 
Motivated by this,
we redefine the mapping function with relational reasoning via graph modeling, and further propose a GCN-based \bf{R}elational \bf{R}easoning \bf{Net}work (RR-Net) in which inter and intra relations are efficiently computed to universally resolve the cross modality mapping problem.
Concretely,
we first construct two kinds of graph, \emph{i.e.}, Intra Graph and Inter Graph, to respectively model intra relations and inter relations.
Then RR-Net updates all the node features and edge features in an iterative manner for learning intra and inter relations simultaneously.
Last, RR-Net outputs the probabilities over the edges which link a pair of heterogeneous instances to estimate the mapping results.
Extensive experiments on three example tasks, \emph{i.e.}, image classification, social recommendation and sound recognition, 
clearly demonstrate the superiority and universality of our proposed model.

\end{abstract}


\IEEEpeerreviewmaketitle

\section{Introduction}

With the explosive growth of multimedia information,
cross modality mapping has attracted much attention in the computer vision community,
the goal of which is to accurately associate a pair of instances from two different modalities.
This research topic has shown great potential in many applications,
such as image caption generation \cite{caption1}\cite{caption2},
visual question answering \cite{VQA1}\cite{VQA2}\cite{VQA},
dimension reduction \cite{dimenReduce}\cite{dimensionTNNLS}, 
domain adaption \cite{domain}, to name a few.
%% figure1
\begin{figure}[!pt]
\begin{center}
\subfigure
{\includegraphics[width=0.47\textwidth]{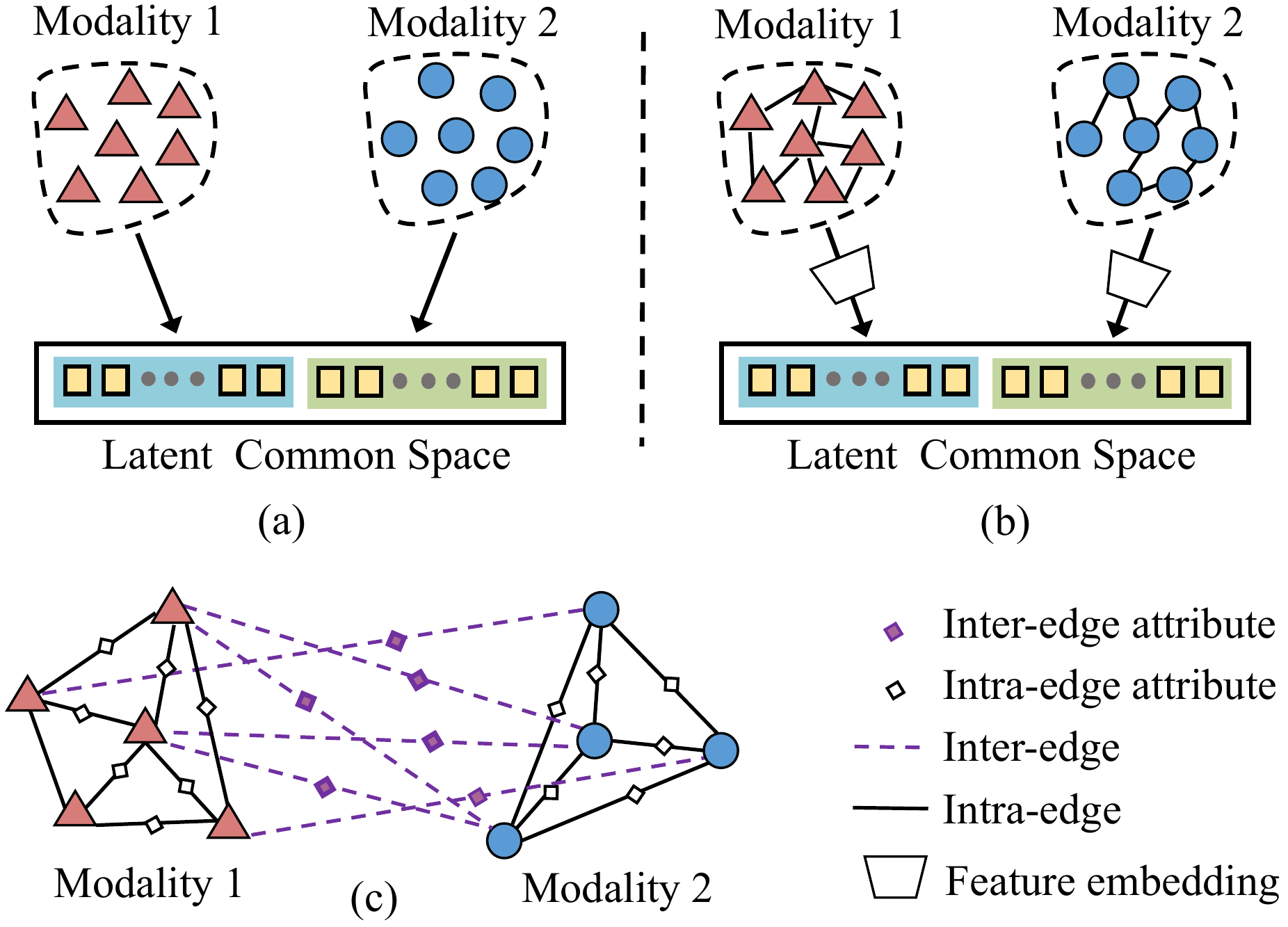}}{}
\captionsetup{margin=2pt,justification=justified}
\caption{The illustration of latent common space based feature embedding methods (a) (b) and
the proposed methods (c) for cross modality mapping.}
\label{fig1}
\end{center}
\end{figure}

Existing cross modality mapping methods rely on the similarity measure between a pair of instances from two modalities to formulate the mapping function.
Therefore, 
how to learn a discriminative feature embedding to represent such similarity for the pair of instances 
(\emph{e.g.}, image \textit{vs} label in image classification area)
plays a crucial role in conventional cross modality mapping task.
In terms of this,
most approaches \cite{matching1}\cite{retrieval2}\cite{retrieval3}\cite{matching2}\cite{uvs}\cite{coor2}\cite{Wsabie}\cite{coor1}\cite{coor3}\cite{retrieval1} learn the embedding by projecting two modalities into a common latent space.
Early studies \cite{matching1}\cite{retrieval2}\cite{retrieval3}\cite{matching2}\cite{uvs}\cite{coor2} only employ linear projection without taking any account of intrinsic relations among the instance
within each single modality (hereinafter called intra relation) or heterogeneous instances from two modalities (hereinafter called inter relation), 
as shown in Fig.~\ref{fig1} (a).
Recent progress \cite{Wsabie}\cite{coor1}\cite{coor3}\cite{retrieval1} turn to incorporate the intra relations for learning the common space embedding by designing two bimodal auto-encoders, as shown in Fig.~\ref{fig1} (b).
However, this line of approaches pays less attention to inter relation, which is critical for supplementing the intra-modality information.
There emerge several other methods \cite{crossmedia}\cite{socialimage} that separately
investigate intra relations and inter relations for learning the embedding in
specific task, \ie, image-text retrieval task.
Nevertheless, viewing that their inter relation is learned without the help of intra relation information from two modalities,
their performance is still heavily limited by the heterogeneous gap of different data modalities.
%Besides, their learned relations are highly dependent on specific data form or distribution}
%Nearly all the prior studies are task-specific, 
%hence their learned relations are highly dependent on specific data form or distribution. 

Above all,
it is critical to explore the intra relations and inter relations simultaneously in a more effective manner for the problem of cross modality mapping.
For any two modality,
we observe that intra relations can be modeled as a structural relationship among instances within a single modality,
while the inter relation can be seen as a reasoning relationship between the pair of instances from two modalities.
Based on this observation,
we naturally leverage graph to well model these two relationships where each instance from two modalities are treated as a node.
Specifically,
two kinds of edges, named as intra-edges and inter-edges, 
are employed to respectively represent intra relations and inter relations.
Thus we redefine the mapping function in this literature via relational reasoning instead of standard similarity measure,
which can be implemented by estimating the existence of the inter-edges.
To the best of our knowledge, unlike task-specific previous arts, 
we are the first attempt to resolve cross modality mapping with relational reasoning and
consider a universal solution that can jointly represent the intra relations and inter relations via graph modeling.

In this work,
inspired by the superiority of \textit{graph convolutional network} (GCN),
we propose a GCN-based \textbf{R}elational \textbf{R}easoning \textbf{Net}work (RR-Net),
 a universal model to resolve the problem of cross modality mapping.
Concretely,
we first construct two kinds of graph: Intra Graph and Inter Graph.
Intuitively,
the former includes two graphs lying in each modality while the latter actually links the instances across two modalities, 
as shown in Fig.~\ref{fig1} (c).
Each Intra Graph takes every instance from the same modality as a node (intra-node) 
and assigns intra-edges via a clustering algorithm, \emph{e.g.}, $K$NN.
As for Inter Graph, the inter-edges
link candidate pairs from two modalities with high initial confidence according to specific task.
On top of the constructed graphs,
our RR-Net first employs an encoder to map the raw features to a desired space,
then simultaneously learn all the node features and edge features 
in an iterative manner via the core component, \emph{i.e.}, 
a relational GCN module implemented by stacking several GCN units.
Finally, RR-Net utilizes a decoder to output the probabilities over the inter-edges to search for the most likely cross modality mapping pairs.
Note that we derive two kinds of GCN units corresponding to Intra Graph and Inter Graph,
\emph{i.e.}, intra GCN unit and inter GCN unit,
each containing one edge convolutional layer (intra- or inter-edge layer) and one node convolutional layer (intra- or inter- node layer).
%All the intra- and inter- node layers are the same for all the GCN units.
The difference between two kinds of GCN units lies in intra- and inter-edge layer, 
which are exploited respectively for learning the intra relations and inter relations.
In particular,
inter-edge layer takes the output of its former intra-node layer as input 
and utilizes a weight matrix as a kernel when performing aggregation of inter-edge features.
%Finally, RR-Net outputs the probabilities over the inter-edges to obtain the reasoning results,
%which is a set of the most likely cross modality mapping pairs.
%Node convolutional layer and intra-edge convolutional layer are extended from \cite{wangtao}. 
%In contrast with \cite{wangtao}, 
%we devise our model on top of  
 %\textit{graph convolutional network} which iteratively performs a feature aggregation from neighbors by message passing, and therefore can express complex interactions among data modalities.

% 第四段，实验验证有效性，并总结创新点 
We conduct extensive experiments on three tasks, 
\emph{i.e.}, sound recognition, image classification and social recommendation,
to verify the universality and effectiveness of our proposed model.
Main contributions of this paper are summarized as follows:

\begin{itemize}
\vspace{0em}
\item We are the first to resolve cross modality mapping with \textit{relational reasoning} and propose a \textit{task-agnostic universal} solution to learn both intra and inter relations simultaneously via graph modeling.
\vspace{0.6em}
\item We propose a GCN-based \textit{Relational Reasoning Network} (RR-Net) to jointly learn all the node and edge features with multiple intra and inter GCN units.
\vspace{0.6em}
\item On several different cross modality mapping tasks with public benchmark datasets,
the proposed RR-Net improves the performance significantly over the state-of-the-art competitors.
 \end{itemize}

%% figure2
\begin{figure*}[!pt]
\begin{center}
\subfigure
{\includegraphics[width=0.95\textwidth]{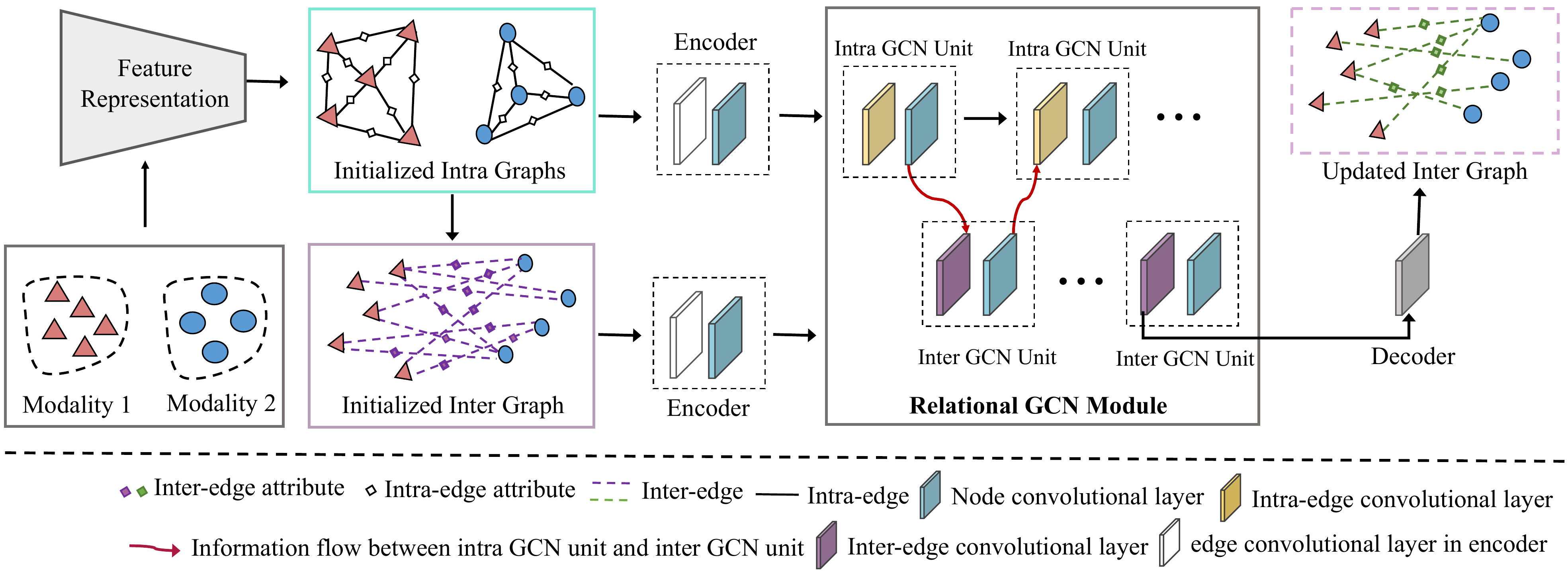}}
%\vspace{-4mm}
\captionsetup{margin=2pt,justification=justified}
\caption{The illustration of overall framework. %It first constructs two initialized Intra Graphs for two different data modalities.
%Then an Inter Graph is initialized on top of the two Intra Graphs.
%After this,
We first construct three graphs, two Intra Graphs and one Inter Graph.
Then our RR-Net utilizes two components: an encoder-decoder and the core Relational GCN Module to model the graphs.
Among which, Relational GCN Module is implemented by stacking two kinds of GCN units, \emph{i.e.}, intra GCN unit and inter GCN unit,
in an iterative manner.
%RR-Net feds all the graphs into an encoder, followed by a relational GCN module.
%Such module
%consists of a set of intra and inter GCN units
%to iteratively update the node and edge features in Intra and Inter Graph, respectively.
Finally,
RR-Net feds the updated inter-edge features into a decoder to produce a set of probabilities to search for the most likely cross modality mapping pairs.
}\label{fig2}
\end{center}
\end{figure*}

\section{Related Work}
In this section,
we first give a briefly review approaches about the cross modality learning in Sec. \ref{cross-modal},
we then introduce studies of relational reasoning and graph neural network that are closely related to this work in Sec.~\ref{relationalReasoning} and Sec.~\ref{gnn}, respectively.

\subsection{Cross Modality Learning}
\label{cross-modal}
Most of the existing cross modality algorithms can 
be classified into two categories,
that is, joint embedding learning and 
coordinated embedding learning.
Below, we briefly review these two categories of approaches.

% 指出我们的方法与别的方法的区别
\subsubsection{Joint Embedding Learning}
This kind of methods embeds data from two modalities together into a common feature space and performs the cross modality similarity measure. 
Studies of \cite{retrieval3}\cite{matching2}\cite{joint2}\cite{joint1}
directly concatenate the features of different modalities to form the common feature space.
Unlike such straightforward method,
some methods \cite{feifeili}\cite{crossmedia}\cite{DBM}\cite{suk}\cite{MDL}\cite{matching1}\cite{retrieval2}\cite{uvs}\cite{WuCVPR} first convert all of the modalities into different representations,
they then concatenate multiple representations together to a joint feature space.
For example,
Ngiam \et. \cite{MDL} stacked several auto-encoders for individually learning the representation of each modality,
then they fused those representations into a common embedding space.
Srivastava \et. \cite{DBM} introduced a multimodal DBMs to fuse multimodal representations.
Following the DBMs,
Suk \et. \cite{suk} utilized milti-modal DBM representation to perform Alzheimer's disease classification from positron emission tomography and magnetic resonance imaging data.
Afterwards,
Wang \et. \cite{wangsubspace} jointly learned several projection matrices to map multi-modal data into a common subspace and measured similarities of different data modality.
Recently,
Wu \et. \cite{WuCVPR} factorized image and its descriptions into different levels to learn a joint space of visual representation and textual semantics.
However,
these approaches only consider the common feature space embedding for each modality,
which ignored the structural interactions between two modalities,
and thus they lack the capacity to represent complicated heterogeneous modality data. 

\subsubsection{Coordinated Embedding Learning}
Instead of projecting the data modalities into a joint space,
the coordinated embedding learning method separately learn the representations for each modality but coordinate them through a constraint,
typically using metric learning \cite{coredularized}, linear transfer \cite{coor1}\cite{retrieval1}\cite{socialimage}, margin ranking loss \cite{Wsabie}\cite{coor3},
pairwise similarity loss \cite{Yu}, \textit{etc}.
For instance,
Andrew \et. \cite{coor1} mapped the multi-modal features into a shared space by learning two linear transfers, and they jointly maximized the correlation
across two modalities to compute their similarities.
WSABIE \cite{Wsabie} and DeVISE \cite{coor3} learned to linearly transform both image and text features into a joint feature space with a margin ranking loss.
Yu \et. \cite{Yu} proposed a dual-path neural network model to learn both image and text feature representations and then learned their correlation with a pairwise similarity loss. 
Although these approaches have achieved great improvement for learning the cross modality mapping,
they merely consider intra relations within single modality and ignore the inter relations between two modalities.
And their learned representations lack distinctiveness and comprehensiveness, thus leading to a severe degradation of performance.
%the robustness and comprehensiveness of the learned representations %may be degraded severely.
In \cite{socialimage},
Huang \et. proposed a joint embedding modal to combine the social relations for representation learning of the multimodal contents. 
However,
this method is particularly designed for social images, which is not suitable for other type of data medias.
In this work,
we represent each data modality as a graph,
and profoundly mine both the intra and inter relations by jointly learning the node features and edge features of different data modalities.

\subsection{Relational Reasoning}
\label{relationalReasoning}
%As a fundamental ability of humans,
Relational reasoning aims to infer about certain relationships between different entities.
It plays an important role in many computer vision tasks such as activity recognition \cite{2014Two}, text detection \cite{textDetection},
video understanding \cite{Temporal},
and visual question answering \cite{reasonVQA}\cite{feng2020scalable}, \textit{etc.}
For learning the intuitive interactions between entities,
many relational approaches \cite{textDetection}\cite{Temporal}\cite{ASN}\cite{feng2020scalable}\cite{DynamicVQA}\cite{2020Better}\cite{2018stmbolic} have been developed.
For example,
Zhang \et. \cite{textDetection} reasoned the linkage relationship between the text components by exploiting a spectral-based graph convolution network.
%Santoro \et. \cite{ASN} devised a dedicated module for computing inter-entity relations.
Zhou \et. \cite{Temporal} designed a temporal
relational network (TRN) to reason about the interactions
between frames of videos in varying scales.
Yi \et. \cite{2018stmbolic} disentangled reasoning from image
and language understanding, by first extracting symbolic
representations from images and text, and then executing
symbolic programs over them.
Gao \et. \cite{DynamicVQA} dynamically fused visual features and question words with intra- and inter-modality information flow, which reasoned their relations by alternatively passing information between and across multi-modalities. 
These relational reasoning methods are commonly split into two stages:
the first one is structured sets of representations extraction, which is intended to correspond to entities from the raw data;
While the second one is how to utilize those representations for reasoning their intrinsic relationships.

Our work mainly focuses on how to utilize the raw representations of data modalities to model both intra and inter relations in cross modality mapping. 
For any two data modalities, 
once we model the intra relations as a structural relationship among instances within a single modality, 
and meanwhile view the inter relations as a reasoning relationship between the pair of instances from two modalities,
then the problem of cross modality mapping can be modeled as a structural and relational reasoning problem.
Intuitively,
we are the first 
attempt to reason about the structural relations within both single and multiple data modalities simultaneously,
which are commonly-existed, important, but ignored by
most existing cross data modality mapping studies. 
With the exploitation of structural relations, 
our model is able to learn the mapping relation between different data modalities more comprehensively.

\subsection{Graph Neural Network}
\label{gnn}
Recently,
graph neural networks \cite{gnn}\cite{gan}\cite{gcn}\cite{gcn2}\cite{gcn3},
especially the graph convolutional network (GCN),
have realized obvious progress because of its expressive power in handing graph relational structures.
It can express complex interactions among data instances by performing feature aggregation from neighbors via message passing.
Studies of \cite{matching5}\cite{penggcn} learned visual relationships among images by applying graph reasoning models.
In \cite{relationGCN},
Michael \et. proposed a relational GCN to learn specific contextual transformation for each relation type. 
Chen \et. \cite{retrieval1} decomposed data modalities into hierarchical semantic levels and generated corresponding embedding 
via a hierarchical graph reasoning network.
More recently,
Wang \et. \cite{faceLinkage} proposed a spectral-based GCN to solve the problem of clustering faces, where the designed GCN can rationally
link different face instances belonging to the same person in complex situations.
Motivated by these studies,
we model the cross modality mapping with relational reasoning via graph modeling, representing each data modality as an Intra Graph and constructing an Inter Graph on top of those Intra graphs.
Upon these graphs,
we further propose a GCN-based relational reasoning Network in which inter and intra relations are efficiently learned to universally resolve the cross modality mapping problem.

%Method
\section{Methodology}
\label{method}
In this section, we first present the overall architecture of
the proposed network in Sec.~\ref{overview}. 
Then we give some preliminaries as well as schemes of graph construction for our method
in Sec.~\ref{graphconstruction}.
Details of the proposed RR-Net are introduced in Sec.~\ref{GCN modeling}. Finally, we describe the loss function for training our RR-Net in Sec.~\ref{loss}.

\subsection{Framework Overview}
\label{overview}
Fig. \ref{fig2} shows the information flow of our proposed method. 
Two Intra Graphs and one Inter Graph are firstly initialized on top of raw representations extracted for each data modality.
Taking these constructed graphs as inputs,
RR-Net then transfers the node and edge attributes of graphs into a latent representations through the encoder module.
Next, RR-Net updates the representations by jointly learning the node features and edge features in an iterative manner via the core relational GCN module.
%RR-Net takes these constructed graphs as input through the encoder
%and then updates them by jointly learning the node features and edge features in an iterative manner via the core relational GCN module.
After that, 
we cast the updated edge features into the decoder to produce a set of probabilities over the inter-edges,
with the goal of obtaining the most likely cross modality mapping pairs.

\subsection{Graph Construction}
\label{graphconstruction}
%\vspace{-1mm}
\subsubsection{Preliminary} Generally, an attributed graph can be represented as
$\mathbb{G}\!=\!(\mathbb{V, E,} \ \nu, \varepsilon)$, where
%in which $\mathbb{V}\!\!=\!\! \{v_{1},...,v_{n}\}$ denotes the node set with $n$ numbers; $\mathbb{E} \subseteq \mathbb{V}\times\mathbb{V}$ denotes the edge set; $\nu = \{\textbf{v}_{i}|\textbf{v}_{i}\in\mathbb{R}^{d_{V}}, i = 1,2,...n\}$ denotes the node attribute set, where ${d_{V}}$ indicates the dimension of node features;  $\varepsilon = \{\textbf{e}_{i}|\textbf{e}_{i}\in\mathbb{R}^{d_{E}}, i=1,2,...|\mathbb{E}|\}$ denotes the edge attribute set, where ${d_{E}}$ indicates the dimension of edge features.
%number of nodes;
\begin{itemize}
\item $\mathbb{V} = \{v_{1},...,v_{n}\}$ denotes the node set, in which $n$ is the number of nodes;
\item $\mathbb{E} \subseteq \mathbb{V}\times\mathbb{V}$ denotes the edge set;
\item $\nu = \{\textbf{v}_{i}|\textbf{v}_{i}\in\mathbb{R}^{d_{V}}, i = 1,2,...n\}$ denotes the node attribute set, where ${d_{V}}$ indicates the dimension of node features;
\item $\varepsilon = \{\textbf{e}_{i}|\textbf{e}_{i}\in\mathbb{R}^{d_{E}}, i=1,2,...|\mathbb{E}|\}$ denotes the edge attribute set, where ${d_{E}}$ indicates the dimension of edge features.
$|\mathbb{E}|$ refers to the number of edges.
 \end{itemize}
Given two modalities, we represent them as two Intra Graphs,
\emph{i.e.},
$\mathbb{G}^{1}=(\mathbb{V}^{1}, \ \mathbb{E}^{1}, \ \nu^{1}, \ \varepsilon^{1})$ and $\mathbb{G}^{2}=(\mathbb{V}^{2}, \ \mathbb{E}^{2}, \ \nu^{2}, \ \varepsilon^{2})$.
On the basis of $\mathbb{G}^{1}$ and $\mathbb{G}^{2}$,
we further construct an Inter Graph $\mathbb{G}^{\mathcal{A}}=(\mathbb{V}^{\mathcal{A}}, \ \mathbb{E}^{\mathcal{A}}, \ \nu^{\mathcal{A}}, \ \varepsilon^{\mathcal{A}})$.
Our goal is to infer a probability set $\mathrm{P}$ for $\mathbb{E}^{\mathcal{A}}$ to predict whether the candidate pairs exist.

\subsubsection{Intra Graph Construction}
Given raw feature representations (extracted from a pre-trained convolutional neural network, \ie, \textit{CNN} networks \cite{resnet}\cite{vgg} for extracting the visual feature, \textit{word2vec} \cite{word2vec} for extracting the text feature, \textit{etc}.) of all instances in each single modality,
we initialize the Intra Graph $\mathbb{G}^{t} (t=1,2)$ by treating each instance $i$ as one intra-node $v_{i}^t$
and then generate intra-edges using task-dependent strategies,
such as $K$NN algorithm in image classification (Sec.~\ref{sec:exp:classification}) and the natural social relation in recommendation system (Sec.~\ref{sec:exp:recommendation}), \etc.
Raw intra-node attributes $\textbf{v}_{i}^t$ are directly derived from the raw feature representations of the instance $i$,
%that is $\textbf{v}_{i}\in\mathbb{R}^{d_{V}}$.
while intra-edge attributes $\textbf{e}_{i}^t$ are initialized by concatenating the attributes of the two associated two intra-nodes,
\emph{i.e.}, $\textbf{e}_{i}^t=\textbf{v}_{s_i}^t\copyright \textbf{v}_{r_i}^t$ where $s_i$ and $r_i$ denote the sender node and the receiver node respectively, and $\copyright$ denotes the concatenate operation.

\subsubsection{Inter Graph Construction}
On top of two Intra Graphs $\mathbb{G}^{1}$ and $\mathbb{G}^{2}$,
we construct an Inter Graph $\mathbb{G}^{\mathcal{A}}$ to model the inter relation between two heterogeneous modalities.
Specifically,
we take all the nodes in two Intra Graphs as the the inter-node set
$\mathbb{V}^{\mathcal{A}} = \mathbb{V}^{1} \cup \mathbb{V}^{2}$.
Each inter-node attribute is represented by inheriting the intra-node set
 $\nu^{1}$ and $\nu^{2}$.
For the inter-edge generation,
a naive way is to build all edges cross the two Intra Graphs $\mathbb{G}^{1}$ and $\mathbb{G}^{2}$.
However, this strategy not only increases the computational cost and memory burden, but also introduces too much noise for inferring the inter relation between the two modalities.
In this paper,
for each inter-node $v_{i}^{\mathcal{A}}$,
we generate only a few inter-edges associated with it with high confidences that is computed according to domain knowledge.
Similarly,
we represent inter-edge attribute $\textbf{e}_{i}^{\mathcal{A}}$ by concatenating the attributes of its associated inter-nodes,
$\textbf{e}_{i}^{\mathcal{A}}=\textbf{v}_{s_i}^{1} \ \copyright \ \textbf{v}_{r_i}^{2}$.
Each inter-edge indicates a candidate mapping between instances of the two modalities, and we develop a deep graph network to learn for selecting reliable inter-edges from the built graphs.

%It is worth noting that the adjacency relationship in Intra Graph is fixed while that in Inter Graph may change from existence to non-existence due to the relational reasoning.

\subsection{RR-Net}
\label{GCN modeling}
Taking the constructed graphs as input,
RR-Net learns to form structured representations for all nodes and edges simultaneously via relational reasoning.
RR-Net contains two modules: the Encoder-Decoder Module and the core Relational GCN Module, 
which are elaborated in Sec.~\ref{sec:method:encoder} and Sec.~\ref{sec:method:GCN}.
%which is illustrated in Fig.~\ref{fig1}.

\subsubsection{Encoder-Decoder Module} \label{sec:method:encoder}
% 第一段，介紹encode-decoder module
The encoder module aims to transfer the edge and node attributes in $\mathbb{G}^{1}$, $\mathbb{G}^{2}$ and
$\mathbb{G}^{\mathcal{A}}$ into latent representations, exploiting two parametric update functions $\Psi^{e}$ and $\Psi^{v}$.
Similar to studies in \cite{retrieval2}\cite{wangtao}, we design the two functions as two \textit{multi-layer-perceptions} (MLPs).
For each graph, the encoder module updates the attributes by applying $\Psi^{v}$ to all nodes and  $\Psi^{e}$ to edges:
%For example, we update $\nu^{\mathcal{A}}$ and $\varepsilon^{\mathcal{A}}$ in $\mathbb{G}^{\mathcal{A}}$ as:
\begin{equation}
\begin{aligned}
\textbf{v}_i^1 \gets \Psi^{v}&(\textbf{v}_i^1),  \ \ \ \
\textbf{e}_i^1 \gets \Psi^{e}(\textbf{e}_i^1), \\
\textbf{v}_i^2 \gets \Psi^{v}&(\textbf{v}_i^2),  \ \ \ \
\textbf{e}_i^2 \gets \Psi^{e}(\textbf{e}_i^2), \\
\textbf{v}_i^{\mathcal{A}} \gets \Psi^{v}&(\textbf{v}_i^{\mathcal{A}}),  \ \ \ \
\textbf{e}_i^{\mathcal{A}} \gets \Psi^{e}(\textbf{e}_i^{\mathcal{A}}). \\
\end{aligned}
\end{equation}
After that, we pass all the graphs to the subsequent relational GCN Module for joint learning of intra and inter relations.

%第二段，介紹decoder module
The decoder module aims to predict a probability $\mathrm{P}\in\mathbb{R}^{|\mathbb{E}^{\mathcal{A}}|}$ over all the inter-edges.
%supposing that the total number of edges in $\mathbb{G}^{\mathcal{A}}$ are $|\mathbb{E}|$.
%Since $\mathrm{P}$ is predicted  the learned edge attribute in inter graph $\mathbb{G}^{\mathcal{A}}$,
Like the encoder, we employ one MLP that is implemented with one parametric update function $\phi$
to transform the inter-edge attribute $\textbf{e}^{\mathcal{A}}$ into a desired space:
\begin{equation}
\begin{aligned}
\textrm{P} = \phi(\textbf{e}^{\mathcal{A}})
\end{aligned}
\end{equation}

\subsubsection{Relational GCN Module} 
\label{sec:method:GCN}
This module is the core component of RR-Net,
aiming to learn all the node features and edge features simultaneously in an iterative manner.
Relational GCN Module is implemented by stacking $L$ copies of two kinds of GCN units, intra-GCN unit and inter-GCN unit, which corresponds to Intra Graph and Inter Graph respectively.
Each GCN unit contains an edge convolutional layer (intra-edge layer or inter-edge layer) and a following node convolutional layer (intra-node layer or inter-node layer).
The intra-node layer and inter-node layer are the same in all the GCN units, while the intra-edge layer and inter-edge layer are derived in a different form for learning the intra relationship and inter relationship respectively,
considering the heterogeneous and interconnected characteristics in cross data modality.

Both kinds of GCN units consist of two steps: message aggregation and message regeneration.
The forward propagation of our model alternatively updates the intra-node attributes and intra-edge attributes through intra GCN unit,
and it then updates the inter-node attributes and inter-edge attributes through inter GCN unit.
Below,
we provide learning process of each unit in detail.

\textbf{(i) Intra-edge convolutional layer}
Taking Intra Graphs $\mathbb{G}^{t} (t=1,2), $ as input, 
the intra-edge layer first employs an aggregation function $\varphi^{e}_{\textrm{intra}}$ that aggregates information of associated nodes for  each intra-edge $\textbf{e}_{i}^{1}$ in $\mathbb{G}^{1}$ and $\textbf{e}_{j}^{2}$ in $\mathbb{G}^{2}$.
Formally,
for $\textbf{e}_{i}^{1}$ and $\textbf{e}_{j}^{2}$, we define its message aggregation as:
%\vspace{-5mm}
\begin{equation}
\begin{aligned}
\hat{\textbf{e}}_{i}^{1} \leftarrow \varphi^{e}_{\textrm{intra}}(\textbf{v}_{s_i}^{1}, \textbf{v}_{r_i}^{1}), \ \
\hat{\textbf{e}}_{j}^{2} \leftarrow \varphi^{e}_{\textrm{intra}}(\textbf{v}_{s_j}^{2}, \textbf{v}_{r_j}^{2})
\end{aligned}
\end{equation}
where $\textbf{v}_{s_i}^{1}\in\nu^{1}, \textbf{v}_{r_i}^{1} \in\nu^{1}$ are the attributes of two connected nodes of edge $\textbf{e}_{i}^{1}$.
Since nodes in the Intra Graph all come from one single data modality, we design the aggregation function $\varphi^{e}_{\textrm{intra}}$ by directly concatenating two node attributes associated with the current edge,
\begin{equation}
\begin{aligned}
\varphi^{e}_{\textrm{intra}}(\textbf{v}_{i},\textbf{v}_{j}) = \textbf{v}_{i} \copyright \textbf{v}_{j}
\end{aligned}
\end{equation}
where $\copyright$ is the concatenate operation of two vectors.
Taking the aggregated information, for example $\hat{\textbf{e}}_{i}^{1}$ and $\hat{\textbf{e}}_{j}^{2}$, as input,
the intra-edge layer adopts a regeneration function $\zeta^{e}_{\textrm{intra}}$ to generate new features and use them to update the intra-edge attributes as:
%\vspace{-3mm}
\begin{equation}
\begin{aligned}
\textbf{e}_{i}^{1} \leftarrow \zeta^{e}_{\textrm{intra}} (\hat{\textbf{e}}_{i}^{1},\textbf{e}_{i}^{1} ), \ \ \
\textbf{e}_{j}^{2} \leftarrow \zeta^{e}_{\textrm{intra}} (\hat{\textbf{e}}_{j}^{2},\textbf{e}_{j}^{2} ) \\
\end{aligned}
\end{equation}
Like \cite{wangtao},
we implement the regeneration function $\zeta^{e}_{\textrm{intra}}$ as an MLP to output an update intra-edge attribute.

\textbf{(ii) Inter-edge convolutional layer}
This layer updates the inter-edge attributes via two functions: an aggregation function $\varphi^{e}_{\textrm{inter}}$
which incorporates its associated inter-node attributes, and an update function $\zeta^{e}_{\textrm{inter}}$ that generates a new
inter-edge attribute.
For each inter-edge $\textbf{e}_{i}^{\mathcal{A}}$ and its associated sender node $\textbf{v}_{s_i}^{\mathcal{A}}$ and receiver node $\textbf{v}_{r_i}^{\mathcal{A}}$,
we define operators in inter-edge layer as:
\begin{equation}
\begin{aligned}
\hat{\textbf{e}}_{i}^{\mathcal{A}} &\leftarrow \varphi^{e}_{\textrm{inter}}(\textbf{v}_{s_i}^{\mathcal{A}}, \textbf{v}_{r_i}^{\mathcal{A}}),
\ \ \\
\textbf{e}_{i}^{\mathcal{A}} &\leftarrow \zeta^{e}_{\textrm{inter}} (\hat{\textbf{e}}_{i}^{\mathcal{A}},\textbf{e}_{i}^{\mathcal{A}} ).
\end{aligned}
\end{equation}
Different from that in the intra-edge layer, we specify the aggregation function $\varphi^{e}_{\textrm{inter}}$ as:
\begin{equation}
\begin{aligned}
\varphi^{e}_{\textrm{inter}}(\textbf{v}_{s_i}^{\mathcal{A}},\textbf{v}_{r_i}^{\mathcal{A}}) = W(\textbf{v}_{s_i}^{\mathcal{A}} \copyright \ \textbf{v}_{r_i}^{\mathcal{A}})
\end{aligned}
\end{equation}
where $W$ is a learnable weight matrix that can be interpreted as a kernel to balance the heterogeneous gap between two modalities.
While for the update function  $\zeta^{e}_{\textrm{inter}}$, we similarly specify it as an MLP that takes the concatenated vector $\hat{\textbf{e}}_{i}^{\mathcal{A}} $\copyright$ \textbf{e}_{i}^{\mathcal{A}}$ as input and outputs an updated inter-edge attribute.

\textbf{(iii) Node convolutional layer}
Following the edge convolutional layer,
the node convolutional layer is used to collect the attributes of all the adjacent edges to the centering node to update their attributes.
In our model,
we design this layer with two functions: an aggregation function $\varphi^{v}$ and an update function  $\zeta^{v}$.
Similar to the edge convolution layer,
for each node $\textbf{v}_{i}^{k}$ in graph $\mathbb{G}^{k} (k\in\{1,2,\mathcal{A}\})$,
we update its attributes as follows:
\begin{equation}
\begin{aligned}
\hat{\textbf{v}_{i}^{k}} &\leftarrow \varphi^{v}\big(\mathbb{E}_{i}^{k} \big), \\
\textbf{v}_{i}^{k} &\leftarrow  \zeta^{v} \big(\hat{\textbf{v}_{i}^{k}}, \textbf{v}_{i}^{k} \big),
\end{aligned}
\end{equation}
where $\mathbb{E}_{i}$ denotes the set of all edges associated with the $\textbf{v}_{i}^{k}$.
Similar to studies \cite{wangtao}\cite{retrieval2},
the aggregation function $\varphi^{v}$ is non-parametric,
and the update function $\zeta^{v}$ is parameterized by an MLP.

\subsection{Loss Function}
\label{loss}
After $L$ iterations of node and edge feature updates, RR-Net outputs the probabilities $\mathrm{P}\!\!\in\!\!\mathbb{R}^{\mathbb{|E^\mathcal{A}|}}$ over the inter-edges from the final decoder module, which is a set of the most likely cross modality mapping pairs.
Then, given the ground-truth mapping $\mathcal{Y}\in \{0,1\}^{\mathbb{|E^\mathcal{A}|}}$ of cross data modality,
we evaluate the difference between the predicted mapping $\mathrm{P}$ and the annotation $\mathcal{Y}$ adopting a $\textit{cross entropy loss}$:
\begin{equation}
\begin{aligned}
\mathcal{L} = -\sum_{i=1}^{|\mathbb{E}^\mathcal{A}|} \big\{ \mathcal{Y}_{i} \ \mathrm{log} (\mathrm{P}_{i}) + (1-\mathcal{Y}_{i}) \ \mathrm{log} (1-\mathrm{P}_{i}) \big\}.
\end{aligned}
\end{equation}

\section{Experiments}
In this section,
we first study key proprieties of the proposed RR-Net on sound recognition task (Sec.~\ref{sec:exp:sound}) and image classification task (Sec.~\ref{sec:exp:classification}).
To examine whether our proposed model can be generalized well in those tasks with the lack of intra relations,
we further verify the proposed model on the social recommendation task (Sec.~\ref{sec:exp:recommendation}).

\subsection{Sound Recognition}
\label{sec:exp:sound}
This task aims to recognize the type of the sound events in an audio streams. In this paper, we verify the effectiveness of RR-Net on learning the mapping between audio and textural data modalities.

\subsubsection{Dataset} 
We evaluate the performance of the proposed RRNet in complex environmental sound recognition task,
on two datasets with different scales: ESC-10 \cite{esc} and ESC-50 \cite{esc} datasets.
The ESC-50 dataset comprehends 2000 audio clips of 5s each.
It equally divides all the clips into fine-grained 50 categories with 5 major groups: animals, natural soundscapes and water sounds, human non-speech sound, interior/domestic sounds, and exterior/urban noises.
%The sound sources in this dataset are from common to quite distinct.
The ESC-10 dataset is a selection of 10 classes from the ESC-50 dataset.
It comprises of 400 audio clips of 5s each.
In our experiments,
we divide the dataset into 5 folds and adopt the leave-one-fold-out evaluations to compute the mean accuracy rate.
For a fair comparison,
we remove completely silent sections in which the value was equal to 0 at the beginning or end of samples in the dataset,
and then convert all sound files to monaural 16-bit WAV files,
following the studies \cite{envnet}\cite{sound1}.

\subsubsection{Implementation Details}
We first construct Intra Graphs, including audio graph $\mathbb{G}^{1}$, textural graph $\mathbb{G}^{2}$,
and Inter Graph $\mathbb{G}^{\mathcal{A}}$,
using schema in Sec.~\ref{method}.
As for $\mathbb{G}^{1}$,
we take each audio as an intra-node
and represent its attribute by  extracting the audio representations of 512 dimension from the baseline model, \ie, EnvNet \cite{envnet}.
Similarly,
$\mathbb{G}^{2}$ takes each text as an intra-node,
it uses the  \textit{word2vec} model to extract text representations of 300 dimension for representing the attribute of each node.
For each intra-node,
its intra-edges are assigned via $K$NN algorithm to search some nearest neighbor nodes.
As for an inter-node in the $\mathbb{G}^{\mathcal{A}}$,
we build its connected edges by selecting top-K nodes which have high initial mapping probabilities according to the baseline model.
We set the number of nearest neighbors of each intra-node in ESC-10
as 5 in $\mathbb{G}^{1}$ and 2 in $\mathbb{G}^{2}$, empirically.
The number of nearest neighbors of each node in ESC-50 is set 
as 10 in $\mathbb{G}^{1}$ and 2 in $\mathbb{G}^{2}$.
Besides,
the top-K parameter in  $\mathbb{G}^{\mathcal{A}}$ is set as 20 on ESC-50, and 10 on ESC-10, respectively.
%% model parameter
We employ one hidden layer in MLP for encoder,
where the number of neurons is empirically set to 16.
In order to better explore the relational reasoning of graph model,
we stack one intra GCN unit and inter GCN unit, thus the total number of GCN units equals to 2.
To train RR-Net,
we use momentum SGD optimizer and set the initial learning rate as 0.01 on ESC-10 and 0.1 on ESC-50, momentum as 0.9 and weight decay as 5e-4.

%% 与STOA对比表
\begin{table}[tp!]
  \centering
  \captionsetup{margin=10pt,justification=justified}
  \caption{Sounds recognition accuracy (\%) on both ESC-10 and ESC-50 dataset. $\S$ means method is training with strong augment and between-class samples. }
%Results of our model are shown in bold font.
\label{soundSOT}
  \begin{tabular}{lcccc}
  \toprule[1pt]
  \specialrule{0em}{2pt}{2pt}
  \multirow{2}*{Methods} & & \multicolumn{3}{c}{Accuracy (\%) on dataset} \\
  \cmidrule(l){3-5}
  \specialrule{0em}{2pt}{2pt}
  & & \bf{ESC-10} & & \bf{ESC-50} \\
  \hline
  \specialrule{0em}{2pt}{2pt}
  M18 \cite{m18} & & 81.8 $\pm$ 0.5 & & 68.5 $\pm$ 0.5\\
  \specialrule{0em}{1pt}{1pt}
  LESM \cite{LESM} & & 93.7 $\pm$ 0.1 & & 79.1 $\pm$ 0.1\\
  \specialrule{0em}{1pt}{1pt}
  DMCU \cite{dmcu} & & 94.6 $\pm$ 0.2 & & 79.8 $\pm$ 0.1\\
  \specialrule{0em}{1pt}{1pt}
  EnvNet \cite{envnet} & & 87.2 $\pm$ 0.4 & & 70.8 $\pm$ 0.1\\
  \specialrule{0em}{1pt}{1pt}
  EnvNet-v2 \cite{bcnet} & & 85.8 $\pm$ 0.8 & & 74.4 $\pm$ 0.3\\
  \specialrule{0em}{1pt}{1pt}
  SoundNet8+SVM \cite{soundnet8} & & 92.2 $\pm$ 0.1 & & 74.2 $\pm$ 0.3\\
  \specialrule{0em}{1pt}{1pt}
  AReN \cite{AReN} & & 93.6 $\pm$ 0.1 & & 75.7 $\pm$ 0.2\\
  \specialrule{0em}{1pt}{1pt}
  EnvNet-v2 $\S$ \cite{bcnet} & & 89.1 $\pm$ 0.6 & & 78.8 $\pm$ 0.3\\
  \specialrule{0em}{1pt}{1pt}
  \bf{RR-Net (Ours)} & & \bf{96.5 $\pm$ 0.1} & & \bf{80.8 $\pm$ 0.1}\\
 \hline
 \specialrule{0em}{2pt}{2pt}
 \textit{Human Performance.} & & 95.7 $\pm$ 0.1 & & 81.3 $\pm$ 0.1\\
 \bottomrule[1pt]
\end{tabular}
%\vspace{-2mm}
%\captionsetup{margin=10pt,justification=justified}
\vspace{-3mm}
\end{table}
\subsubsection{Comparison with State-of-the-arts}
\label{sound results}
We compare our RR-Net with 8 state-of-the-art sounds recognition methods, including 
M18 \cite{m18}, LESM \cite{LESM}, DMCU \cite{dmcu}, EnvNet \cite{envnet},
EnvNet-v2 \cite{bcnet}, SoundNet8+SVM \cite{soundnet8} AReN \cite{AReN} and EnvNet-v2+strong augment \cite{bcnet} (\wrt, EnvNet-v2 $\S$).
Tab. \ref{soundSOT} reports their accuracy results on both ESC-10 and ESC-50 datasets. 
We observe that our proposed model achieves the best performance with 96.5\% and 80.8\% accuracy on the ESC-10 and ESC-50 datasets, respectively.
It has a significant improvement of 2.0\% on the  ESC-10 dataset and 1.0\% on the ESC-50 dataset, compared with the second best method DMCU~\cite{dmcu}.
Note that for the EnvNet-v2 $\S$,
despite of authors pre-process the training data using strong augments and between classes audio samples,
accuracies of this method are obviously lower than us by 7.4\% and 2.0\% on  ESC-10 and  ESC-50 dataset respectively.
It is worth noting that when comparing to the human performance,
our model promotes the recognition accuracy by 0.8\% on ESC-10 dataset.
Since the accuracy achieved by human is already quite high,
thus the improvement of our model is indeed significant.
More specifically, 
by comparing the baseline model EnvNet \cite{envnet},
our model yields an accuracy boost around 9.0\% and 10.0\% on ESC-10 and ESC-50 respectively, as shown in Tab. \ref{soundSOT}.
These results obviously illustrate the effectiveness of our RR-Net for solving mapping among audio \textit{vs} textural modality.

\subsection{Image Classification}
\label{sec:exp:classification}
Previous section clearly illustrates the effectiveness of our RR-Net on audio and textual modality mapping. 
In this section,
we further verify the universality and effectiveness of our model on learning the mapping between image and textual modality.
Taking image classification as an example,
we are not to obtain state-of-the-art results on this task,
but to give room for potential accuracy and robustness improvements in exploring a universal cross modality mapping model.
Below,
utilizing different networks, including ResNet18 \cite{resnet}, ResNet50 \cite{resnet}, and MobileNetV2 \cite{mobile} as baseline model respectively,  we reproduce them for image classification at first,
and then we evaluate our model on top of those baselines for universality illustration.

\subsubsection{Dataset} We adopt two different scales of image classification datasets: 
CIFAR-10 \cite{cifar} and CIFAR-100  \cite{cifar} for our evaluation.
The CIFAR-10 consists of 60, 000 32x32 color images belonging
to 10 categories, with 6,000 images for each category.
This dataset is split into 50, 000 training images and 10,000 test images. 
The CIFAR-100 is just like the CIFAR-10, except it has 100 classes containing 600 images each. There are 500 training images and 100 testing images per class. The 100 classes in the CIFAR-100 are grouped into 20 superclasses. Each image comes with a "fine" label (the class to which it belongs) and a "coarse" label (the superclass to which it belongs).
Data augmentation strategy includes random crop
and random flipping is used during training, following in studies \cite{resnet}\cite{cifar-2}.

\subsubsection{Implementation Details}\label{details} 
Similar to the previous task,
we also construct two Intra Graphs: image graph $\mathbb{G}^{1}$
and text graph $\mathbb{G}^{2}$, and one Inter Graph $\mathbb{G}^{\mathcal{A}}$.
The Intra Graphs respectively take each image and each text as an intra-node.
$\mathbb{G}^{1}$ builds attribute of each intra-node by extracting image features of 512 dimension from the baseline model.
While $\mathbb{G}^{2}$ adopts the \textit{word2vec} model to extract the textural features of 300 dimension for representing attribute of each node.
For each intra-node,
we assign its intra-edge using $K$NN clustering algorithm for searching its connected nodes.
As for an inter-node in $\mathbb{G}^{\mathcal{A}}$,
we build its connected edges by selecting top-K nodes with high initial probabilities according to the baseline model.
We empirically set the number of nearest neighbors for each intra-node in CIFAR-10  as 10 in $\mathbb{G}^{1}$ and 2 in $\mathbb{G}^{2}$,
while 20 and 3 numbers for the CIFAR-100 settings.
The top-K parameter in  $\mathbb{G}^{\mathcal{A}}$ is set as 15 on CIFAR-100, and 10 on CIFAR-10, respectively.
%% model parameter
Similar to previous tasks,
we employ one hidden layer in MLP for encoder,
where the number of neurons is empirically set to 16 for image classification.
we stack two intra GCN units and three inter GCN units, thus the total number of GCN units equals to 5.
Finally,
we train the baselines and RR-Net using the SGD optimizer with initial learning rate 0.01, momentum 0.9, weight decay 5e-4, shuffling the training samples.

\subsubsection{Comparison with Baselines}
Tab. \ref{classification} presents comparison results of top-1 accuracy between our model and the baseline models.
Obviously,
we receive great improvements over different baseline models on both datasets.
In particularly,
RR-Net improves  the classification accuracy over ResNet18,
ResNet50 and MobileNetV2 by 2.59\%, 1.68\%, 1.79\% 
on CIFAR-10 dataset, and 2.01\%, 1.23\%, 3.23\% on CIFAR-100 dataset, correspondingly. 
These results clearly demonstrate the effectiveness of the proposed model for the  image-textural modality mapping.
Tab. \ref{classification} illustrates that better performance
of image classification can be achieved by using better backbones such as ResNet-50, MoblieNetV2,
but thanks to the relational reasoning ability of RR-Net,
our model further improves their performance with a large margin.
Besides,
it also can be seen that 
using different baselines that have different feature representing performance for initializing graphs in our model,
RR-Net consistently improves their accuracy, 
demonstrating the generalization ability of our proposed model. 
 
%% 与STOA对比表
\begin{table}[!t]
\begin{center}
\setlength\tabcolsep{2.5mm}
\captionsetup{margin=10pt,justification=justified}
\caption{Image classification on top-1 accuracy (\%) of different baseline methods and RR-Net on  CIFAR-10 and CIFAR-100 datasets.}\label{classification}
\begin{tabular}{lcccc}
\toprule[1pt]
\specialrule{0em}{1pt}{1pt}
\multirow{2}*{Baseline} & \multirow{2}*{\textbf{RR-Net}} & \multicolumn{3}{c}{Top-1 Accuracy (\%) on}  \\ 
\cmidrule(l){3-5}
& & \textbf{CIFAR-10} & & \textbf{CIFAR-100} \\
\hline
\specialrule{0em}{1pt}{1pt}
ResNet18 & \xmark & 87.04 && 62.55 \\
\specialrule{0em}{1pt}{1pt}
ResNet18 & \cmark & 89.63 && 64.56 \\ 
\hline
\specialrule{0em}{1pt}{1pt}
ResNet50 & \xmark & 90.78 && 72.53 \\
\specialrule{0em}{1pt}{1pt}
ResNet50 & \cmark & 92.46 && 73.76 \\ 
\hline
\specialrule{0em}{1pt}{1pt}
MobileNetV2 & \xmark & 92.36 && 66.61 \\
\specialrule{0em}{1pt}{1pt}
MobileNetV2 & \cmark & 94.15 && 69.84 \\
\bottomrule[1pt]
\end{tabular}
\end{center}
\vspace{-4mm}
\end{table} 

\subsection{Social Recommendation}
\label{sec:exp:recommendation}
We further verify the generalization of our model 
on Social recommendation which aims to provide personalized item suggestions to each user, according to the user-item rating records.
In this task,
social network for users are explicitly provided via the rating records,
but no any relation information existed between items.
Thus,
we evaluate our model on this task under lacking intra relations.
%To this objective,
%our model solves it in a transductive semi-supervised edge regression manner.

\subsubsection{Dataset}  
We evaluate the performance of our model in social recommendation task, adopting two public dataset: Filmtrust \cite{filmtrust} and Ciao \footnote{
http://www.public.asu.edu/~jtang20/datasetcode/truststudy.htm\label{ciao}}.
Details of these two datasets are presented in Tab. \ref{socialdata}.
As for Ciao\textsuperscript{\ref{ciao}},
we filter out all the user nodes and item nodes whose length of id is larger that 99999, since they are the confident unreliable id records.
With those filter nodes,
we naturally remove their connected social links and rating links.
To ensure the high generalization of our model,
We randomly spit each data set into training, valuation and testing data set.
The final performance is gained by meaning results of five times jointly training and testing the corresponding data set.

\subsubsection{Evaluation metrics}
we evaluate our model by two widely used metrics, namely mean absolute error (MAE) and root mean square error (RMSE).
Formally, these metrics are defined as:
\begin{equation}
\begin{aligned}
\textrm{MAE} &= \frac{\sum_{u,j}|\hat r_{u,j}-r_{u,j}|}{N} \\
\textrm{RMSE}& = \sqrt{\frac{\sum_{u,j}(\hat r_{u,j}-r_{u,j})^{2}}{N}}
\end{aligned}
\end{equation}
where $r_{u,j}\in \mathcal{R}$ is the rating record for user $i$ and item $j$.
$\hat r_{u,j}$ is the predicted rating of user $i$ on item $j$, and $N$ is the number of rating records.
Smaller values of MAE and RMSE indicate better performance.

\begin{table}[!t]
\centering
 \setlength\tabcolsep{2.5mm}
 \captionsetup{margin=10pt,justification=justified}
 \caption{Dataset statistics. The rating information exist on $\mathcal{R}$ 
and social information are available on $\mathcal{S}$.}
\label{socialdata}
\begin{tabular}{ccccc}
   \toprule[1pt]
  \specialrule{0em}{2pt}{2pt}
  Dataset & user & item & rating \ ($\mathcal{R}$) & social \ ($\mathcal{S}$) \\
  \hline
  \hline
  \specialrule{0em}{2pt}{2pt}
  \bf{Ciao} & 6,052 & 5,042 & 117,369 & 88,244 \\
  \hline
  \specialrule{0em}{2pt}{2pt}
   \bf{FilmTrust} & 1,508 & 2,071 & 35,497 & 1,853 \\
  \bottomrule[1pt]
\end{tabular}
\end{table}

%% 与STOA对比表
\begin{table}[!t]
  \begin{center}
  \captionsetup{margin=10pt,justification=justified}
  \caption{Comparison results of MAE and RMSE values on FilmTrust and Ciao dataset.
Our method consistently achieves better performance than the previous state-of-the-art approaches.}\label{socialSTOA}
  \setlength\tabcolsep{1.7mm}
  \begin{tabular}{lcccc}
  \toprule[1pt]
  \specialrule{0em}{2pt}{2pt}
   \multirow{2}*{\bf{Methods}} & \multicolumn{2}{c}{\bf{FilmTrust}\cite{filmtrust}} &  \multicolumn{2}{c}{\bf{Ciao\textsuperscript{\ref{ciao}}}}\\
   \cmidrule(l){2-3} \cmidrule(l){4-5}
   \specialrule{0em}{2pt}{2pt}
   &  MAE $\downarrow$ & RMSE $\downarrow$ &  MAE $\downarrow$ & RMSE $\downarrow$ \\
   \hline
   \specialrule{0em}{1pt}{1pt}
   SoReg \cite{soreg} &  0.674 & 0.878 &  1.306 & 1.547 \\
   SVD++ \cite{svd} &  0.659 & 0.846 &  0.844 & 1.188 \\
    \specialrule{0em}{1pt}{1pt}
   SocialMF \cite{socialmf} & 0.638 & 0.837 &  0.946 & 1.254 \\
    \specialrule{0em}{1pt}{1pt}
   TrustMF \cite{trustmf} &  0.650 & 0.833 &  0.937 & 1.212 \\
    TrustSVD \cite{trustsvd} &  0.649 & 0.832 &  0.925 & 1.202 \\
    \specialrule{0em}{1pt}{1pt}
   LightGCN \cite{lightgcn} &  0.669 & 0.893 &  0.796 & 1.037 \\
    \specialrule{0em}{1pt}{1pt}
   \textbf{RR-Net (Ours)} &  \bf{0.646} & \bf{0.824} &   \textbf{0.825}  & \textbf{1.050} \\
    \bottomrule[1pt]
\end{tabular}
\end{center}
\vspace{-5mm}
%\captionsetup{margin=10pt,justification=justified}
\end{table}

\begin{figure*}[!t]
\begin{center}
%{\includegraphics[width=0.24\textwidth]{figgg/esc-layer.png}}
{\includegraphics[width=0.265\textwidth,height=0.205\textwidth]{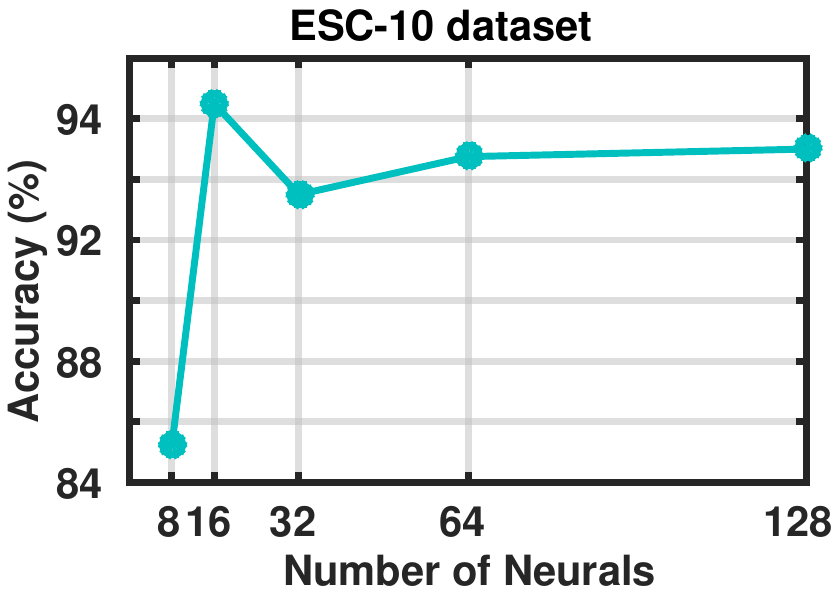}}
{\includegraphics[width=0.26\textwidth,height=0.20\textwidth]{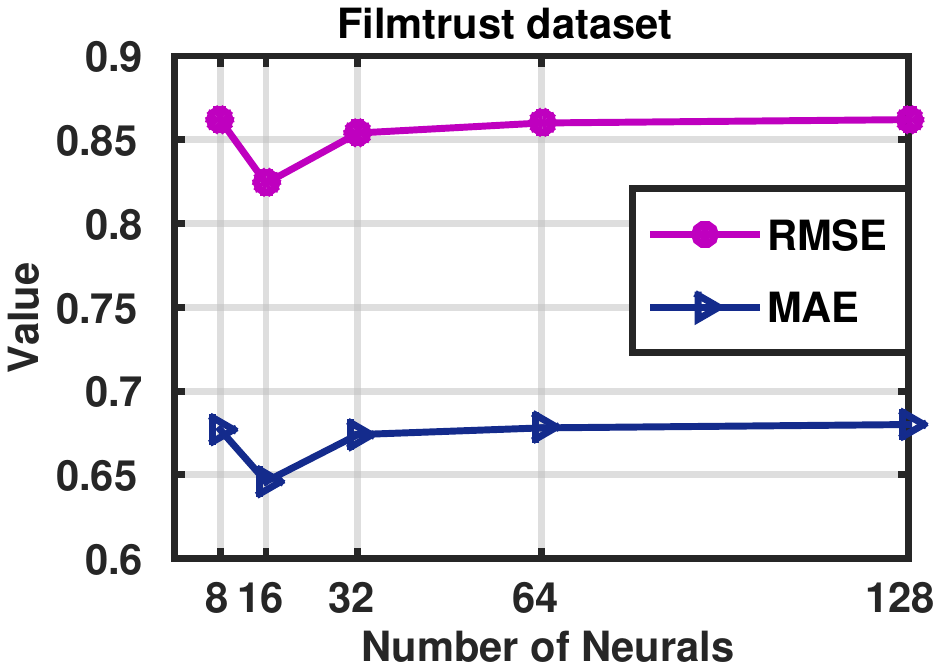}}
{\includegraphics[width=0.26\textwidth,height=0.20\textwidth]{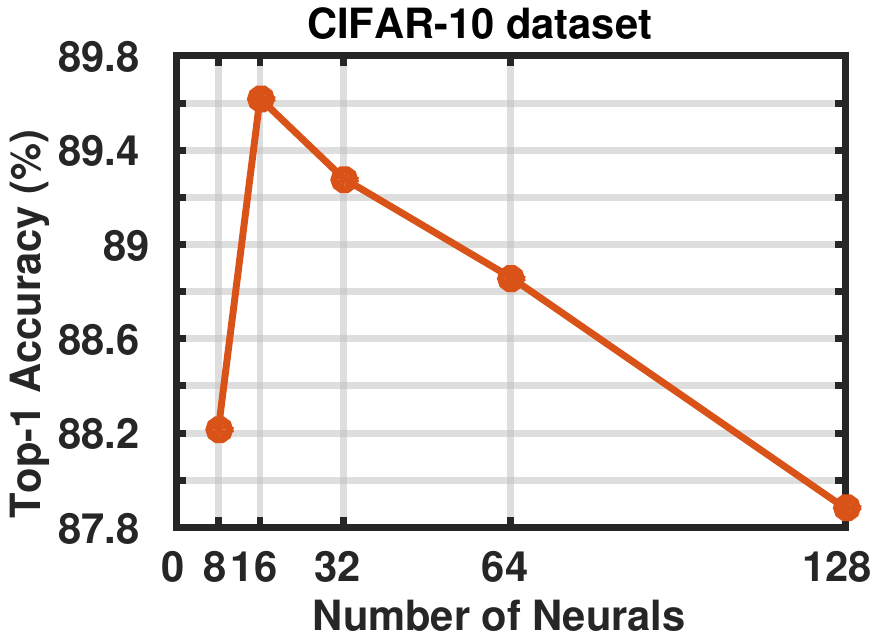}}
\vspace{-2mm}  
\captionsetup{margin=20pt,justification=justified}
\caption{Comparison results of different number of neurons in latent space (\wrt. the number of MLP neural between encoder and decoder) over  ESC-10, Filmtrust and CIFAR-10 dataset, respectively.}\label{encoder-decoder}
\end{center}
\end{figure*}

\begin{figure*}[!t]
\begin{center}
%{\includegraphics[width=0.24\textwidth]{figgg/esc-layer.png}}
{\includegraphics[width=0.26\textwidth]{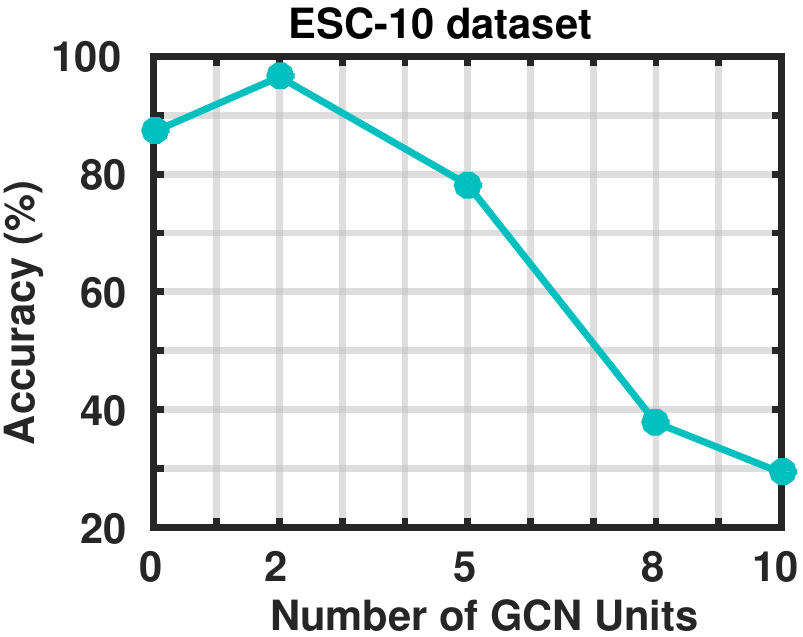}}
{\includegraphics[width=0.268\textwidth]{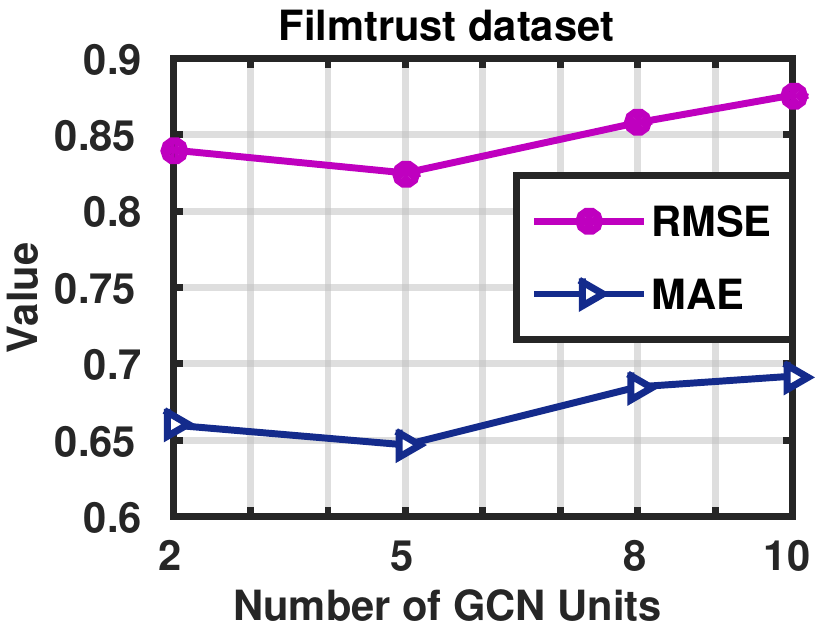}}
{\includegraphics[width=0.26\textwidth]{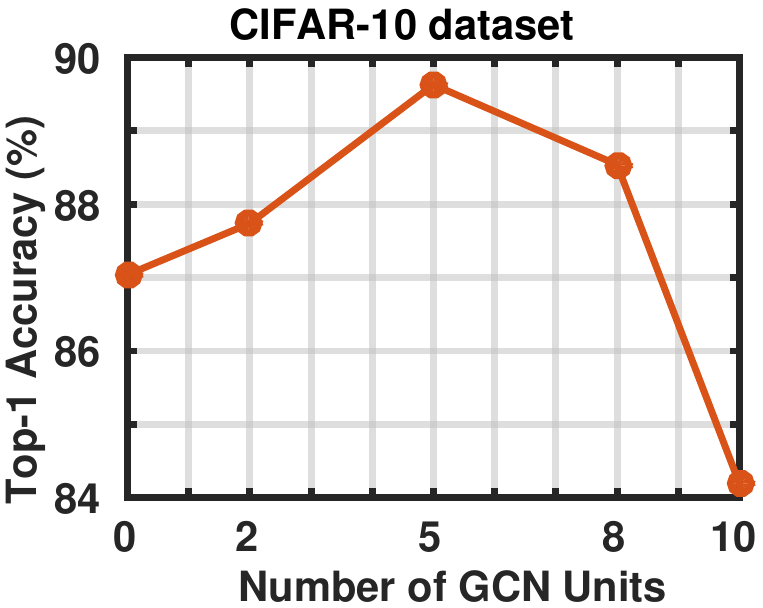}}
\vspace{-2mm}   
\captionsetup{margin=20pt,justification=justified}
\caption{Comparison resluts of different number of GCN units in our RR-Net over  ESC-10, Filmtrust and CIFAR-10 dataset, respectively.}\label{unit}
\end{center}
\vspace{-4mm}
\end{figure*}

\subsubsection{Implementation Details}
In order to predict the potential rating between specific user and item, 
we also construct two Intra Graphs: user graph $\mathbb{G}^{1}$ and item graph $\mathbb{G}^{2}$, and one Inter Graph $\mathbb{G}^{\mathcal{A}}$ on two dataset respectively.
For the intra-node in the Intra Graph,
$\mathbb{G}^{1}$ and $\mathbb{G}^{2}$ build it with each user and each item in the dataset, respectively.
Then we generate the embedding of each user and item from a uniform distribution within [0,1), with dimension of 128,
which is used to represent attribute of each intra node in $\mathbb{G}^{1}$ and $\mathbb{G}^{2}$, respectively.
The intra-edge in $\mathbb{G}^{1}$ according to users’ social relation 
$\mathcal{S}$ provided by dataset.
Since items in dataset are represented by a set of id records,
no any relations between items can be used because of the ambiguous relationship.
Thus,
we build $\mathbb{G}^{2}$
without using any intro edges.
As for inter-graph $\mathbb{G}^{\mathcal{A}}$,
we build the inter edge by linking 
each user to all the items for FilmTrust.
But on the Ciao\textsuperscript{\ref{ciao}},
we build inter edge by linking to the users appeared in its rating links $\mathcal{R}$.
Similar to the sound recognition task, 
one hidden layer in MLP for encoder with the number of 16 are utilized in RR-Net.
Two intra GCN units and three inter GCN units are stacked for exploring the relational reasoning in our relational GCN module.
Finally,
we train our model using the SGD optimizer with initial learning rate  0.01, momentum 0.9, weight decay 5e-4, and 500 epochs on both the two dataset until convergence.

\subsubsection{Comparison with State-of-the-arts}
We compare our model with 6 state-of-the-art social recommendation methods, including
SoReg \cite{soreg},
SVD++ \cite{svd},
TrustSVD \cite{trustsvd},
SocialMF \cite{socialmf},
TrustMF \cite{trustmf}
and LightGCN \cite{lightgcn}.
For a fair comparison,
we reproduce these methods in type of rating prediction and report their results in Tab. \ref{socialSTOA}.
From the results,
it can be seen that our model greatly outperforms most state-of-the-arts over the FilmTrust,
despite lacking of intra relations in item graph. 
Particularly,
by comparing the recent method LightGCN \cite{lightgcn},
RR-Net reduces the recognition error  by around 6\% and 
2\% in terms of RMSE and MAE on the FilmTrust, respectively.
While on the Ciao,
although no intra relation can be exploited during our relational reasoning,
our RR-Net still achieves comparable performance without using any prior knowledge for the intra relations reasoning.
This speaks well that our model is effective and generalized for learning the mapping between different data modalities.

\subsection{Universality Analysis}
%Previous experiments obviously demonstrate the effectiveness and generalization ability of the presented RR-Net for tackling different types of cross modality mapping problems.
%Here,
To further prove the universality of the presented RR-Net,
we also analyze some common characteristics over the above mentioned cross modality mapping tasks.
%We also analyze some common characteristics
%for further proving the universality of the presented RR-Net.
Firstly,
RR-Net is trained with nearly the same learning rate 0.01 for different tasks,
which is stable and robust for training under different data domains.
Secondly,
we vary different number of neurons (\wrt,  the number of MLP neurons before output from decoder) in the encoder-decoder module,
and find that our model can achieve better performance using same length of neurons for different tasks.
Fig. \ref{encoder-decoder} gives the corresponding comparison result curves.
We can see that RR-Net consistently performs the best under the same setting,
\emph{i.e.}, the number of neurons equals to 16 on ESC-10, FilmTrust and CIFAR-10 dataset.
This proves that our RR-Net is not affected by the dimension of latent space. 
\begin{table}[!t]
\renewcommand\arraystretch{1.2}
\centering
\small
\setlength\tabcolsep{1.2mm}
\caption{Accuracy for top-K nearest neighbor nodes for one inter-node in Inter Graph, by setting different values for K.}
\label{soundkg}
\begin{tabular}{c|lccccc}
\toprule[1pt]
\multirow{3}*{\textbf{ESC-50}} & \multicolumn{6}{c}{\textbf{\normalsize Impact of $k_{\mathcal{A}}$}} \\
\cline{2-7}
& Size of $k_{\mathcal{A}}$ & 5 & 15 & \bf{20} & 40 & 50 \\
& Accuracy (\%) & 79.4 & 79.7 & \bf{80.8} & 79.6 & 76.8  \\
\hline
\specialrule{0em}{2pt}{2pt}
\multirow{2}*{\textbf{CIFAR-100}} & Size of $k_{\mathcal{A}}$ & 10 & \bf{15} & 20 & 50 & 100 \\
 & Accuracy (\%) & 63.24 & \bf{64.56} & 63.46 & 60.09 & 50.73 \\
\bottomrule[1pt]
\end{tabular}
\vspace{-4mm}
\end{table}
Moreover,
we also notice that RR-Net is not sensitive to the total number of GCN units when performing the relational reasoning,
as illustrating in Fig. \ref{unit}.
From this figure,
it can be seen that our model performs the best by setting 
the total number of GCN units as 2, 5, 5 for ESC-10, FilmTrust and CIFAR-10 dataset, respectively.
This illustrates our model is effective by setting the GCN unit within a range of [2, 5] for different tasks.
Similar situation can be found in Tab.~\ref{soundkg}.
We adopt different number of candidate inter-edges for one inter-node to construct Inter Graph on both sound recognition and image classification task.
%we variant the number of neurals (\wrt,  the dimension before output from decoder) for the latent space.
%Best performance can be reached by setting neural numbers as 16 on three datasets,
Resulting in Tab.~\ref{soundkg} shows that our model reaches better performance with the number of top-K nearest neighbors ranging from 15 to 20,
which further illustrates the strengthen universality of our model for  cross modality mapping leaning.
Interestingly,
it also can be seen that performance would be consistently decreased when building inter-edge by fully connecting intra-nodes in two Intra Graphs.
This mainly because of too much noise are introduced in the Inter Graph,
which limits the reasoning ability of RR-Net for inferring the inter relation between the two modalities.
On the contrary,
our model exhibits best performance with several high confident inter-edges in the Inter Graph.

\section{Conclusion}
In this paper,
we resolve the cross modality mapping problem with relational reasoning via graph modeling and propose a universal
RR-Net to learn both intra relations and inter relations simultaneously.
Specifically,
we first construct Intra Graph and Inter Graph.
On top of the constructed graphs,
RR-Net mainly takes advantage of Relational GCN module to update the node features and edge features in an iterative manner,
which is implemented by stacking multipleGCN units. 
%Two kinds of GCN units with different edge convolutional layers are derived for intra and inter relation reasoning.
Extensive experiments on different types of cross modality mapping clearly demonstrate the
superiority and universality of our proposed RR-Net.

\ifCLASSOPTIONcaptionsoff
  \newpage
\fi

\bibliographystyle{IEEEtran}
\bibliography{egbib}

%\vfill


%% insert the photos of all authors
\vspace{-10mm}

\end{document}